\documentclass{article}

\newcommand{\cjy}[1]{{\color{black}#1}}
\newcommand{\jy}[1]{{\color{black}#1}}

\usepackage{spconf}
 
\usepackage{graphicx}
\usepackage{textcomp}

\usepackage{enumitem}
\usepackage{algorithmicx}
\usepackage{algorithm} 
\usepackage{threeparttable}
\usepackage{multirow}
\usepackage{subfigure}
\usepackage{booktabs}
\usepackage[noend]{algpseudocode}
\usepackage{hyperref}
\usepackage[table,xcdraw]{xcolor}
\usepackage{tablefootnote}
\usepackage{stfloats}
\usepackage{float}

\usepackage{amsmath,amssymb,amsfonts}

\title{Vision, Deduction and Alignment: An Empirical Study on Multi-modal Knowledge Graph Alignment}
\makeatletter
\def\name#1{\gdef\@name{#1\\}}
\makeatother
\name{\em{Yangning Li$^{1,4*}$,
    Jiaoyan Chen$^{2*}$,
    Yinghui Li$^{1}$,
    Yuejia Xiang$^{3}$,} 
    Xi Chen$^{3}$,
    Hai-Tao Zheng$^{1,4\dag}$
\thanks{* Equal contribution.}
\thanks{\dag \ Corresponding author. (E-mail: zheng.haitao@sz.tsinghua.edu.cn)}
}

\address{$^{1}$Shenzhen International Graduate School, Tsinghua University \\
      $^{2}$Department of Computer Science, The University of Manchester \\
      $^{3}$Tencent, $^{4}$Peng Cheng Laboratory}
\begin{document}
\ninept
\maketitle
\begin{abstract}
 \cjy{Entity alignment (EA) for knowledge graphs (KGs) plays a critical role in knowledge engineering. Existing EA methods mostly focus on utilizing the graph structures and entity attributes (including literals), but ignore images that are common in modern multi-modal KGs.
In this study we first constructed Multi-OpenEA --- eight large-scale, image-equipped EA benchmarks,
and then evaluated some existing embedding-based methods 
for utilizing images.
In view of the complementary nature of visual modal information and logical deduction, we further developed a new multi-modal EA method named LODEME using logical deduction and multi-modal KG embedding, with state-of-the-art performance achieved on Multi-OpenEA and other existing multi-modal EA benchmarks.
}
%

\end{abstract}

\section{Introduction}
Entity alignment (EA) aims to find out entities referring to the same real-world object from different knowledge graphs (KGs). It plays a critical role in \jy{KGs application and NLP field~\cite{DBLP:journals/corr/abs-2211-04215, dong2021survey}.}
With the development of representation learning~\cite{DBLP:conf/ijcnn/LiWLZS22}, embedding-based EA methods have taken the dominant position \jy{in recent research} \cite{sun13benchmarking,zhang2020industry,zhang2021comprehensive}.
Despite \jy{quite a few positive results have been} achieved, these methods still have some limitations. They over-rely on the structure and literal (entity names) information~\cite{DBLP:conf/sigir/LiLHYS022,DBLP:journals/corr/abs-2207-08087}. In dealing with long-tailed entities which have \jy{little} structural information \cjy{as the evidence for alignment}, \jy{the performance is relatively poor}.

We find entities of many modern KGs are associated with images which are often known as the \emph{visual modality}; e.g.,
each entity in DBpedia is associated with $6.2$ images in average.
In building domain KGs, such as goods KG in e-commerce \cite{luo2020alicoco}, it's also common to add images.
These images often contain strong evidences for checking the equivalence of entities.
Different from the literals, the images are less heterogeneous w.r.t. feature learning, even when they come from \jy{different} KGs. However, as far as we know, the entity images are ignored by most existing EA methods and benchmarks. 

Although, some recent approaches \cite{liu2021visual,xu2022relation} utilize images for EA, they cannot effectively utilize multiple images but can only handle single images. Worse, they fail to perform logical deduction in the inference phase like some conventional non-embedding based methods, which has proved to be quite useful by recent study\cite{sun13benchmarking}. Especially in multi-modal scenarios, visual modal information and logical deduction capabilities are highly complementary~\cite{DBLP:journals/patterns/LiuLTLZ22}: for long-tailed entities (containing less structural information) that don't allow effective deduction, multiple images can provide high-quality visual alignment signals; while for non-long-tailed entities, especially some central entities with high degrees (e.g., \textit{US, China}), they have plenty of related images and no single representative image can denote this entity alone. This is when visual modals fail, but logical inference can carry out effective inconsistency repair and holistic estimation with the assurance of extensive symbolic knowledge~\cite{DBLP:journals/corr/abs-2211-10997}.

Meanwhile, there is a shortage of benchmarks for evaluating EA methods that \jy{consider} the visual modality.
The benchmarks by \cite{liu2021visual} have limited scales and low image coverage (roughly 70\% of the entities have images and each entity has at most one image), and thus they do not match the real-word scenarios and hinder the development of multi-modal EA.

In this work we \textit{first constructed a series of large-scale multi-modal EA benchmarks named Multi-OpenEA} \jy{with a high ratio of image equipped entities and multiple images per entity, based on the OpenEA benchmarks} \cite{sun13benchmarking}.
We then \textit{evaluated four competitive unimodal embedding-based EA models for thier compatibility with visual modalities}, including BootEA~\cite{sun2018bootstrapping}, MultiKE, RDGCN~\cite{wu2019relation} and IMUSE~\cite{he2019unsupervised}.
As a result, the visual modality improves the performance of all these models with an average of 12\% Hit@1 rise, demonstrating that visual modality has general and significant validity for existing embedding-based methods.

In view of the complementary nature of visual modal information and logical deduction, we finally propose a self-supervised EA model LODEME that iteratively performs \textbf{LO}gical \textbf{DE}duction and \textbf{M}ulti-modal \textbf{E}mbedding.
We also developed a structure-aware attention mechanism such that the entity embedding\jy{s} can incorporate multiple images with different emphases.
LODEME outperforms the modified embedding-based models and other multi-modal EA methods, with Hit@1 exceeding 95\% on the Multi-OpenEA benchmarks. With this empirical study\footnote{The source codes and benchmarks will be public.}, we also conducted exhaustive ablation experiments.

\section{Multi-OpenEA Benchmarks}
\cjy{We proposed a generic multi-modal EA benchmarks construction process and constructed new multi-modal EA benchmarks based on the eight existing OpenEA benchmarks\footnote{\url{https://github.com/nju-websoft/OpenEA}} by adding multiple images to each entity.}
\cjy{The construction includes the following three steps:}

\noindent\textbf{Step 1. Entity Name Acquisition.} 
\cjy{The entities in the original OpenEA benchmarks are from either DBpedia (DBP) or Wikidata (WD).
For a DBpedia entity, its name is extracted from its Uniform Resource Identifier (URI) via regular expressions. For a Wikidata entity, its named is accessed via Wikidata SPARQL endpoint\footnote{\url{https://query.wikidata.org/}} using the built-in property \textit{rdfs:label}.}

\noindent\textbf{Step 2. \cjy{Entity Image} Acquisition.} We search the entity name through \jy{the} Google search engine \jy{and get} ten most relevant images for each entity. 
\jy{If the search engines returns less than ten images, we get the remaining}
 entity images from the original KGs (DBpedia and Wikidata) by SPARQL query. 

\noindent\textbf{Step 3. Image Sampling.} We randomly sampled 
\cjy{$3$} images for each entity without considering search engine ranking which may cause bias.

\label{sec:benchmarks}
\begin{table}[h]
\centering
\caption{\cjy{Our Multi-OpenEA benchmarks vs the existing multi-modal EA benchmarks. \jy{Ours} have larger scale (\textbf{\#Entity}), more entities associated with images (\textbf{Coverage}), and more images per entity (\textbf{Ratio}).}
}
\scalebox{0.7}{
\renewcommand\arraystretch{1}
\setlength\tabcolsep{2.5pt}
\begin{tabular}{llccccc}
\toprule
\textbf{Benchmark}              & \textbf{KGs} & \multicolumn{1}{l}{\textbf{\#Entity}} & \multicolumn{1}{l}{\textbf{\#Images}} & \multicolumn{1}{l}{\textbf{Ratio}} & \multicolumn{1}{l}{\textbf{Coverage}} & \multicolumn{1}{l}{\textbf{Similarity}} \\ \midrule
\multirow{2}{*}{\shortstack{FB15K-DB15K\\ \cite{chen2020mmea}} } & FB15K  & 14,951                        & 13,444   & 0.899      & 90.0\%    & \multirow{2}{*}{-}                      \\ \cmidrule{2-6} 
                     & DB15K  & 12,842                        & 12,837    & 0.999                    & 99.9\%                         \\ \midrule
\multirow{2}{*}{\shortstack{DBP-WD(norm)\\\cite{liu2021visual}}}    & DBP   & 15,000                        & 8,517 & 0.517                    & 57.1\%      & \multirow{2}{*}{-}                    \\ \cmidrule{2-6} 
                     & WD   & 15,000                        & 8,791    & 0.586                    & 58.6\%                         \\ \midrule\midrule
\multirow{2}{*}{EN-FR-15K-V1}    & EN15K(V1)    & 15,000                        & 44,657  & 2.977                       & 99.7\% & \multirow{2}{*}{0.757}                       \\ \cmidrule{2-6} 
                     & FR15K(V1)   & 15,000                        & 42,286   & 2.819                     & 94.5\%                         \\ \midrule
                     \multirow{2}{*}{EN-FR-15K-V2}    & EN15K(V2)    & 15,000                        & 44,932  & 2.995                       & 99.9\% & \multirow{2}{*}{0.767}                       \\ \cmidrule{2-6} 
                     & FR15K(V2)   & 15,000                        & 42,622   & 2.841                     & 94.5\%                         \\ \midrule
                     \multirow{2}{*}{EN-FR-100K-V1}    & EN100K(V1)    & 100,000                        & 296,934  & 2.969                       & 99.6\%  & \multirow{2}{*}{0.751}                       \\ \cmidrule{2-6} 
                     & FR100K(V1)   & 100,000                        & 280288   & 2.803                     & 94.1\%                         \\ \midrule
                     \multirow{2}{*}{EN-FR-100K-V2}    & EN100K(V2)    & 100,000                        & 299,403  & 2.994                       & 99.9\%        & \multirow{2}{*}{0.752}                       \\ \cmidrule{2-6} 
                     & FR100K(V2)   & 100,000                        & 282,063   & 2.821                     & 94.4\%                         \\ \midrule
                    \multirow{2}{*}{D-W-15K-V1}    & DBP15K(V1)    & 15,000                        & 44,776  & 2.985                       & 99.8\%          & \multirow{2}{*}{0.829}                       \\ \cmidrule{2-6} 
                     & WD15K(V1)   & 15,000                        & 44,823   & 2.988                     & 99.8\%                         \\ \midrule
                    \multirow{2}{*}{D-W-15K-V2}    & DBP15K(V2)    & 15,000                        & 44,911  & 2.994                       & 99.9\%         & \multirow{2}{*}{0.820}                       \\ \cmidrule{2-6} 
                     & WD15K(V2)   & 15,000                        & 44,945   & 2.996                     & 99.9\%                         \\ \midrule
                    \multirow{2}{*}{D-W-100K-V1}    & DBP100K(V1)    & 100,000                         & 296.749  & 2.9867                      & 99.5\%        & \multirow{2}{*}{0.833}                       \\ \cmidrule{2-6} 
                     & WD100K(V1)   & 100,000                         & 297,354   & 2.974                     & 99.6\%                         \\ \midrule
                    \multirow{2}{*}{D-W-100K-V2}    & DBP100K(V2)   & 100,000                        & 299,338     & 2.993                   & 99.9\%    & \multirow{2}{*}{0.832}                       \\ \cmidrule{2-6} 
                     & WD100K(V2)   & 100,000                        & 299,607    & 2.996            & 99.9\%                        \\ \bottomrule
\end{tabular}}
\label{tab:dataset_compare}
\end{table}

\cjy{We constructed $8$ Multi-OpenEA benchmarks}
which \cjy{are named as X-Y-Z. X denotes the KG source with the values of \{D-W, EN-FR\}. D-W is a cross-KGs version, whose KGs are derived from DBpedia and Wikidata respectively. EN-FR is a cross-lingual version, whose KGs are derived from English version and French version of DBpedia respectively.; Y denotes the KG scale (entity number) with values of \{15K, 100K\}; Z has the values of V1 and V2, which correspond two different versions of the original OpenEA benchmarks with different average relation degrees. Specifically, average relation degrees of V2 is roughly twice that of V1.}
%
%
In Table \ref{tab:dataset_compare}, we compare \cjy{Multi-OpenEA benchmarks with the existing multi-modal} benchmarks used in MMEA \cite{chen2020mmea} and EVA \cite{liu2021visual}. 
\cjy{Due to larger scales, more entities associated with images and higher image number per entity, the Multi-OpenEA benchmarks are more consistent with practical scenarios and more challenging for the methods.} The image similarities between alignment entity pairs in the last column illustrate that the visual modality provides a high-quality alignment signal, which are calculated by the cosine similarity of the CLIP embeddings.

To further evaluate the quality of the automatically acquired images, we randomly selected 3000 images corresponding to 1000 entities and employed 3 annotators to judge whether the correspondence between images and entities was correct. The average accuracy rate was 88.1\% with an inter-annotator agreement of 0.853, measured by Fleiss’s Kappa \cite{fleiss1971measuring}.

\section{Methods}
\cjy{In this section, we introduce how to extend some typical embedding-based EA methods as well as our LODEME for utilizing \jy{the visual modality}.}

\subsection{Modified Embedding-based Models}
\label{sec:base}
\cjy{We consider BootEA, MultiKE, RDGCN and IMUSE for \jy{extension} since they are very recent and typical embedding-based \jy{methods}, often achieving state-of-the-art performance on many EA benchmarks without images  \cite{sun13benchmarking,zhang2020industry}.}
The modified models \cjy{are named by adding the suffix ``-V''.} 

\noindent\textbf{BootEA-V.} BootEA is a semi-supervised method \cjy{based on some translation-based KG embedding models such as TransE.
In BootEA-V, We used a multi-modal translation-based model \cite{mousselly2018multimodal}, whose energy function leverages both visual and structural information.}

\noindent\textbf{MultiKE-V.} MultiKE utilizes multi-view learning to encode entities based on the views of names, relations and attributes. We used the multi-modal pre-training model CLIP \cite{radford2learning} with fixed parameters as the feature extractor for images. Then we leveraged wide-used PCA \cite{dunteman1989principal} algorithm to reduce the dimensionality of the image embedding to \jy{that of} the name embedding. Finally, MultiKE-V treats visual information as \jy{an additional} view of the entity, just like entity name.

\noindent\textbf{RDGCN-V.} RDGCN is a \cjy{Graph Convolutional Network-based models} which uses Word2Vec embeddings \cite{le2014distributed} of entity names as the initial weights of entities. Instead, \cjy{RDGCN-V sets} the weights of entities to the average of entity name embeddings and image embeddings.

\noindent\textbf{IMUSE-V.} IMUSE \cjy{is an interactive method which calculates} entity alignment and attribute alignment alternately. \cjy{The entity similarity is calculated based on the literal (attribute value) similarity.}
\cjy{IMUSE-V uses the entity image as a special literal} and refines the entity similarity with the cosine similarity of the image embeddings.

\subsection{LODEME}
\label{sec:prase}
\subsubsection{Architecture and Pipeline}

\begin{figure}[]
    \centering
    \scalebox{0.75}{\includegraphics[width = 0.98 \linewidth]{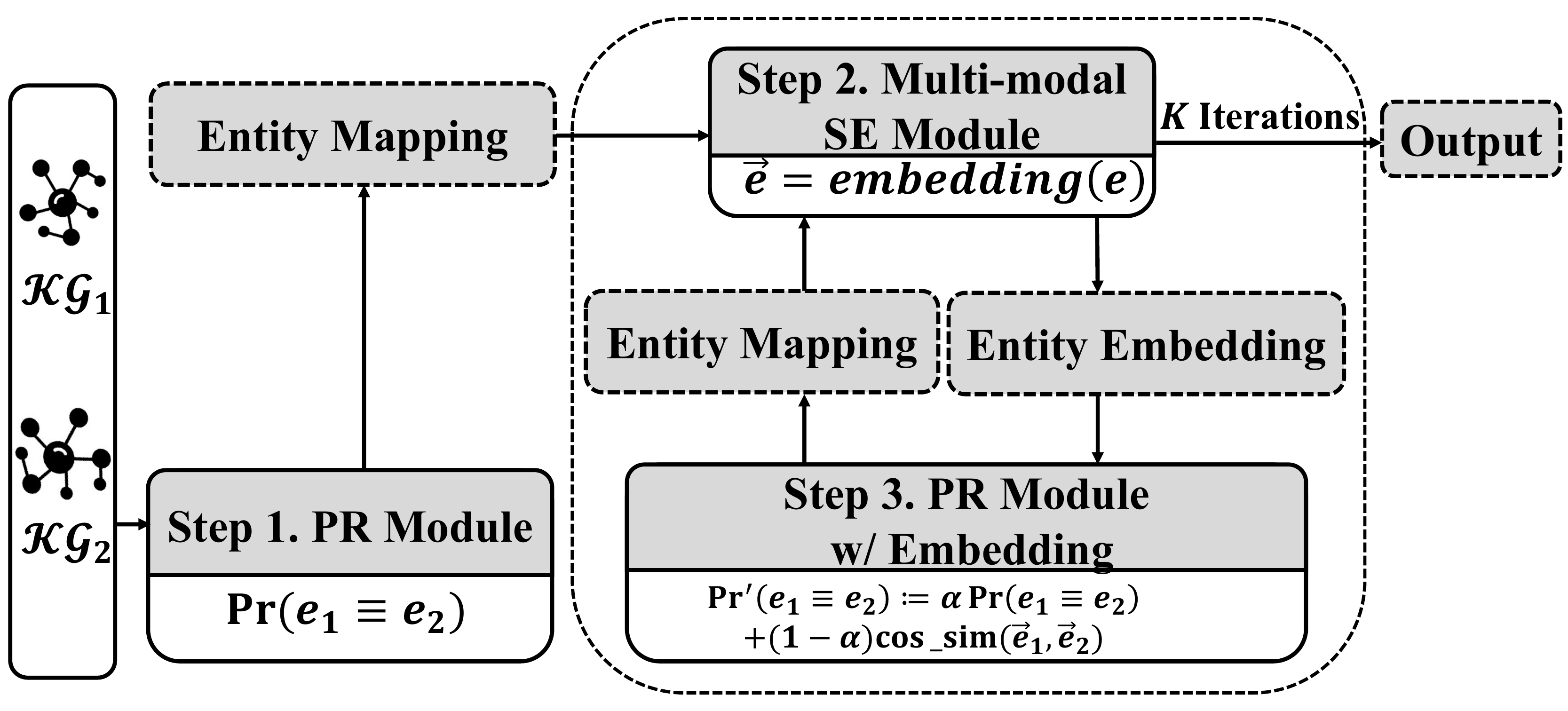}}
    \caption{The Architecture of LODEME.}
    \label{fig:model}
\end{figure}

\cjy{As shown in Figure \ref{fig:model},
LODEME} consists of a probabilistic reasoning (PR) module and a multi-modal semantic embedding (SE) module. 
\cjy{It runs Step 1 only once as an initialization, and then runs \cjy{S}teps 2 and 3 alternately by several times.}

\noindent\textbf{Step 1. \cjy{Run} PR Module.} The PR module \jy{gets the mappings by iterative name matching and probabilistic reasoning (see Section \ref{sec:pr} Probabilistic Reasoning Module).}·

\noindent\textbf{Step 2. \cjy{Run} Multi-modal SE Module.} The entity mappings \cjy{from} the PR module \cjy{are} used as \cjy{the seed mappings (input)} of the SE module for training. \cjy{The SE module encodes multi-modal information and outputs entity embeddings (see} Section \ref{sec:multimodal_se} Multi-Modal Embedding Module).

\noindent\textbf{Step 3. \cjy{Re-run 
PR Module.}} 
Unlike \jy{Step 1}, the \jy{reasoning} in this step, which takes into account the similarity of entity embeddings, enabling the PR module to possess \cjy{multi-modal} information and thus reason more accurately.

\subsubsection{Probabilistic Reasoning Module}
\label{sec:pr}
The PR module in LODEME is developed based on the conventional \cjy{reasoning-based} system PARIS \cite{suchanek2011paris}. First, 
PARIS pre-calculates the functionalities $\mathrm{F}(r)$ and inverse functionality $\mathrm{F}^{-1}(r)$ of the relations in KGs, which are used to portray the uniqueness of the corresponding tail (head) entities for a certain relation given the head (tail) entity. 

Then, PARIS mines the initial entity mapping by lexical matching, \cjy{and updates the probabilities of entity mappings $\mathrm{Pr}\left(e_{1} \equiv e_{2}\right)$ 
and relation mappings (subordination)} $\mathrm{P}\left(r_1 \subseteq r_2\right)$ alternatively based on relation functionalities. Notice that these two probabilities are interdependent, so PARIS self-iterates multiple times until convergence.
Finally, the PARIS outputs \cjy{entity mappings with probabilities to the SE module}. We highly recommend referring to \cite{suchanek2011paris} for more details.

In \cjy{S}tep 1, the PR module just uses original PARIS, while in \cjy{S}tep 3, the PR module is adapted with some changes. It refines the probability estimation of entity alignment by multi-modal entity embedding similarity. The revised PR module \cjy{to exploit not only} logical reasoning but also the \cjy{KGs' multi-modal information encoded by} the SE module. Specifically, the \cjy{equivalence} probability of entities is redefined as $ \mathrm{Pr}^{'}\left(e_{1} \equiv e_{2}\right) :=\alpha \mathrm{Pr}\left(e_{1} \equiv e_{2}\right) + \left(1-\alpha \right) \mathrm{cos\_sim}\left(\vec{\mathbf{e}}_{1}, \vec{\mathbf{e}}_{2}  \right)$, where the first term is the original probability estimate in PARIS and the second term represents the cosine similarity of the corresponding multi-modal entity embedding\cjy{s}.

\subsubsection{Multi-Modal Embedding \cjy{Module}}
\label{sec:multimodal_se}
The high-quality 
\cjy{seed mappings unsupervisedly generated by the PR module are fed to the SE module where the KG modalities are encoded in the following way}.

\noindent\textbf{Structure embedding.}
To capture the structural information of the KGs, we utilize the Graph Convolution Network (GCN) in \cite{DBLP:conf/iclr/KipfW17}. We used the output of the last layer of a three-layer
GCN as the graph structure embedding $\mathbf{F}_{G}$.

\noindent\textbf{Relation and attribution embedding.}
The correlation between relations \cjy{and between attributions can also be used as important information for finding equivalent entities}. Inspired by \cite{yang2019aligning}, we encode the relation and attribute information of entities separately as:
\begin{equation}
\begin{aligned}
\mathbf{F}_{R}=\mathbf{W}_{R} \cdot \mathbf{R}+\mathbf{b}_{R}, \mathbf{F}_{A}=\mathbf{W}_{A} \cdot \mathbf{A}+\mathbf{b}_{A},
\end{aligned}
\end{equation}
where $\mathbf{R} \in {\mathbb{R}}^{m \times r}$ and $\mathbf{A} \in {\mathbb{R}}^{m \times a}$ are count matrices for relations and attributes respectively. $m$, $r$, $a$ \cjy{denote} the number of entities, relations and attributes. $\mathbf{W}_{R}$ and $\mathbf{W}_{A}$ are trainable weight matrices. $ \mathbf{b}_{R}$ and $ \mathbf{b}_{A}$ are trainable bias matrices.

\noindent\textbf{Entity name embedding.}
We used the mean pooling of token representations from the final layer of Multi-lingual BERT \cite{pires2019multilingual} to encode entity name $
    \mathbf{F}_{N}=\mathbf{W}_{N} \cdot \mathrm{M\mbox{-}BERT}
    (\mathrm{name})+\mathbf{b}_{N},$
where $\mathbf{W}_{N}$ and $\mathbf{b}_{N}$ are trainable matrices.

\noindent\textbf{Image embedding.} 
We \cjy{used CLIP for image feature extraction}. Since an entity corresponds to multiple images, we developed a \textit{structure-aware attention mechanism} to make the model place different emphasis on \cjy{different images in different entities:}
\begin{equation}
\begin{split}
    \mathbf{F}_{I}=\sum_{i=0}^{n}\left [\frac{e^{\mathbf{F}_{G}\cdot\mathbf{F}_{I_i}}}{\sum_{j=0}^{n}e^{\mathbf{F}_{G}\cdot\mathbf{F}_{I_j}} }\cdot \mathbf{F}_{I_i} \right ], \mathbf{F}_{I_i}=\mathbf{W}_{I} \cdot \mathrm{CLIP}({I_i})+\mathbf{b}_{I},
\end{split}
\end{equation}
\cjy{We concatenate the embeddings of all the modalities with trainable weights 
as the entity embedding:
\begin{equation}
\small
\vec{\mathbf{e}}=
\bigoplus_{\mathbf{F} \in \left\{\mathbf{F}_{G}, \mathbf{F}_{R}, \mathbf{F}_{A}, \mathbf{F}_{N}, \mathbf{F}_{I} \right\}}\left[\mathrm{softmax}(\mathbf{W}) \cdot \mathbf{F}\right],
\end{equation}
where $\mathbf{W}$ denotes the weights of modalities. 
} 

For training, we employ the widely used \cite{wang2018cross,wu2019jointly} margin-based alignment loss function $L$, which expects greater similarity between positive pairs in alignment mappings and less similarity between negative pairs generated. Formally, $L$ is defined as follows:
\begin{equation}
L=\sum_{(\vec{\mathbf{e_1}}, \vec{\mathbf{e_2}}) \in \mathbb{P}} \sum_{\left(\vec{\mathbf{e_1^{\prime}}}, \vec{\mathbf{e_2^{\prime}}}\right) \in \mathbb{N}} \max \left\{0, sim(\vec{\mathbf{e_1^{\prime}}},\vec{\mathbf{e_2^{\prime}}})-sim(\vec{\mathbf{e_1}},\vec{\mathbf{e_2}})+\gamma\right\}
\end{equation}
where $sim(\cdot)$ is the cosine similarity and $\gamma\textgreater 0$ is a margin hyper-parameter. $\mathbb{P}$ are the entity pairs contained in the alignment mappings provided by PR module. We generate negative pairs $\mathbb{N}$ by hard negative sampling for stronger distinguishing capability. Given a positive pair $(e_1,e_2)$, we choose $K$ nearest entities of $e_1$ ($e_2$) in another KG to replace $e_2$ ($e_1$) to form hard negative pairs.

For inference, We simply used greedy nearest neighbour search and cross-domain similarity local scaling (CSLS) \cite{lample2018word} over entity embeddings as the final result.

\section{Experiments}
\subsection{Experimental Setup}
 For the original embedding-based models, \cjy{we follow the setting in \cite{sun13benchmarking}}. 
We report the higher of the results \jy{we reproduce and the results reported} in \cite{sun13benchmarking}.\footnote{Except for BootEA on D-W-15K-V2, where the claimed result far exceeds the reproduced result.}
We also \cjy{use the latest multi-modal EA method EVA~\cite{liu2021visual} and homochronous MSNEA \cite{xu2022relation} as baselines}.
5-fold cross-validation is used in performance measurement. For fairness, all models use the Multi-lingual BERT and CLIP as encoders for text and images. 

\begin{table}[ht]
\centering
\renewcommand\arraystretch{1}
\setlength\tabcolsep{3pt}
\caption{Overall results on the Multi-OpenEA benchmark\cjy{s}.
}
\scalebox{0.55}{
\begin{tabular}{|l|l|ccc|ccc|ccc|ccc|}
\hline
\rowcolor[HTML]{EFEFEF} 
\multicolumn{2}{|l|}{\cellcolor[HTML]{EFEFEF}}                   & \multicolumn{3}{c|}{\cellcolor[HTML]{EFEFEF}15K-V1}                                                                                                       & \multicolumn{3}{c|}{\cellcolor[HTML]{EFEFEF}15K-V2}                                                                                                       & \multicolumn{3}{c|}{\cellcolor[HTML]{EFEFEF}100K-V1}                                                                                                      & \multicolumn{3}{c|}{\cellcolor[HTML]{EFEFEF}100K-V2}                                                                                                      \\ \cline{3-5} \cline{6-8} \cline{9-11} \cline{12-14}
\rowcolor[HTML]{EFEFEF} 
\multicolumn{2}{|l|}{\multirow{-2}{*}{\cellcolor[HTML]{EFEFEF}}} & \multicolumn{1}{l}{\cellcolor[HTML]{EFEFEF}Hit@1} & \multicolumn{1}{l}{\cellcolor[HTML]{EFEFEF}Hit@5} & \multicolumn{1}{l|}{\cellcolor[HTML]{EFEFEF}MRR} & \multicolumn{1}{l}{\cellcolor[HTML]{EFEFEF}Hit@1} & \multicolumn{1}{l}{\cellcolor[HTML]{EFEFEF}Hit@5} & \multicolumn{1}{l|}{\cellcolor[HTML]{EFEFEF}MRR} & \multicolumn{1}{l}{\cellcolor[HTML]{EFEFEF}Hit@1} & \multicolumn{1}{l}{\cellcolor[HTML]{EFEFEF}Hit@5} & \multicolumn{1}{l|}{\cellcolor[HTML]{EFEFEF}MRR} & \multicolumn{1}{l}{\cellcolor[HTML]{EFEFEF}Hit@1} & \multicolumn{1}{l}{\cellcolor[HTML]{EFEFEF}Hit@5} & \multicolumn{1}{l|}{\cellcolor[HTML]{EFEFEF}MRR} \\ \hline
                                    & BootEA                    & 0.618                                            & 0.795                                               & 0.697                                             & \underline{0.488}                                           & 0.704                                              & 0.584                                           & 0.516                                              & 0.685                                               & 0.594                                            & 0.766                                              & 0.892                                               & 0.822                                            \\ 
                                    & BootEA-V              & 0.730                                              & 0.901                                               & 0.805                                             & \underline{0.728}                                              & 0.926                                               & 0.814                                            & 0.643                                              & 0.837                                               & 0.730                                            & 0.830                                              & 0.937                                               & 0.866                                            \\  
                                    & MultiKE                   & \underline{0.426}                                              & 0.513                                               & 0.471                                             & 0.561                                              & 0.723                                               & 0.636                                            & \underline{0.291}                                              & 0.352                                               & 0.324                                            & \underline{0.327}                                              & 0.410                                               & 0.371                                            \\ 
                                    & MultiKE-V             & \underline{0.737}                                             & 0.771                                               & 0.754                                             & 0.727                                              & 0.765                                               & 0.746                                            & \underline{0.743}                                             & 0.766                                               & 0.755                                            & \underline{0.687}                                              & 0.727                                               & 0.707                                            \\ 
                                    & RDGCN                   & 0.561                                              & 0.714                                               & 0.722                                             & 0.640                                              & 0.777                                               & 0.702                                            & 0.362                                              & 0.485                                               & 0.420                                            & 0.421                                              & 0.528                                               & 0.473                                            \\
                                    & RDGCN-V                   & 0.683                                              & 0.800                                               & 0.736                                             & 0.686                                              & 0.817                                               & 0.744                                            & 0.537                                              & 0.656                                               & 0.592                                            & 0.489                                              & 0.704                                               & 0.584                                            \\
                                    & IMUSE                   & 0.327                                              & 0.523                                               & 0.419                                             & 0.581                                              & 0.778                                               & 0.671                                        & 0.276                                              & 0.437                                               & 0.355                    & 0.431                                              & 0.631                                               & 0.525                                                                        \\
                                    & IMUSE-V                   & 0.404                                              & 0.593                                               & 0.492                                             & 0.606                                              & 0.806                                               & 0.696                           & 0.351                                              & 0.521                                               & 0.432                     & 0.494                                              & 0.701                                               & 0.590                                                                                    \\ \cline{2-14}
                                    & PARIS                   & 0.734                                              & -                                               & -                                             & 0.840                                              & -                                               & -                                            & 0.667                                              & -                                               & -                                            & 0.795                                              & -                                               & -                                             \\
                                    & MSNEA                   & 0.962                                              & 0.988                                               & 0.973                                             & 0.971                                              & 0.974                                               & 0.989                                            & 0.946                                              & 0.957                                               & 0.952                                            & 0.982                                              & 0.988                                               & 0.989                                            \\
                                                                        & EVA                  & 0.971                                              & 0.989                                               & 0.978                                             & 0.990                                              & 0.998                                               & 0.994                                            & 0.968                                              & 0.989                                               & 0.976                                            & 0.991                                              & 0.998                                               & 0.994                                            \\
\multirow{-11}{*}{\rotatebox{90}{D-W}}                  & LODEME               & \textbf{0.991}                                              & 0.998                                              & 0.994                                             & \textbf{0.996}                                            & 1.000                                               & 0.998                                            & \textbf{0.973}                                              & 0.992                                               & 0.973                                            & \textbf{0.994}                                              & 0.999                                               & 0.996 \\ \hline \hline     
                                    & BootEA                    & \underline{0.507}                                              & 0.718                                               & 0.603                                             & \underline{0.660}                                              & 0.850                                               & 0.745                                            & 0.389                                              & 0.561                                               & 0.474                                            & 0.640                                              & 0.806                                               & 0.716                                            \\ 
                                    & BootEA-V              & \underline{0.717}                                              & 0.918                                               & 0.806                                             & \underline{0.807}                                              & 0.898                                               & 0.845                                            & 0.509                                              & 0.732                                               & 0.611                                            & 0.706                                              & 0.786                                               & 0.744                                            \\  
                                    & MultiKE                   & 0.796                                              & 0.876                                               & 0.834                                             & 0.868                                              & 0.920                                               & 0.892                                            & 0.629                                              & 0.680                                               & 0.655                                            & 0.642                                              & 0.696                                               & 0.670                                            \\ 
                                    & MultiKE-V             & 0.916                                              & 0.961                                               & 0.937                                             & 0.936                                              & 0.964                                               & 0.949                                            & 0.661                                              & 0.739                                               & 0.699                                                     & 0.675                                              & 0.725                                              & 0.705                                            \\ 
                                    & RDGCN                   & 0.817                                              & 0.909                                               & 0.858                                             & 0.847                                              & 0.919                                               & 0.880                                            & 0.640                                              & 0.732                                               & 0.683                                            & 0.715                                              & 0.787                                               & 0.748                                            \\
                                    & RDGCN-V                   & 0.862                                              & 0.927                                               & 0.891                                             & 0.900                                              & 0.956                                               & 0.925                                            & 0.730                                             & 0.928                                               & 0.816                                            & 0.779                                              & 0.934                                               & 0.846                                            \\
                                    & IMUSE                   & 0.569                                              & 0.717                                               & 0.639                                             & 0.607                                              & 0.760                                               & 0.678                                            & \underline{0.439}                                              & 0.546                                               & 0.492                                            & \underline{0.461}                                              & 0.605                                               & 0.529                                            \\
                                    & IMUSE-V                   & 0.663                                              & 0.776                                               & 0.715                                             & 0.736                                              & 0.862                                               & 0.792                                            & \underline{0.568}                                              & 0.6710                                               & 0.618                                            & \underline{0.570}                                              & 0.700                                               & 0.631                                            \\ \cline{2-14}
                                    & PARIS                   & 0.903                                              & -                                               & -                                             & 0.934                                              & -                                               & -                                            & 0.848                                              & -                                               & -                                            & 0.881                                              & -                                               & -                                            \\
                                    & MSNEA                   & 0.978                                              & 0.990                                               & 0.981                                             & 0.982                                              & 0.998                                               & 0.990                                            & 0.927                                              & 0.941                                               & 0.938                                            & 0.965                                              & 0.985                                               & 0.973                                            \\
                                                                        & EVA                   & 0.982                                              & 0.996                                               & 0.988                                             & 0.993                                              & 1.000                                               & 0.996                                            & 0.940                                              & 0.968                                               & 0.950                                            & 0.971                                              & 0.995                                               & 0.980                                            \\
\multirow{-11}{*}{\rotatebox{90}{EN-FR}}  & LODEME               & \textbf{0.989}                                              & 0.997                                               & 0.992                                             & \textbf{0.997}                                              & 1.000                                               & 0.998                                            & \textbf{0.966}                                              & 0.983                                               & 0.972                                            & \textbf{0.978}                                              & 0.996                                               & 0.985   \\ \hline
\end{tabular}}
\vspace{-0.2cm}
\label{tab:main}
\end{table}

\subsection{Main Experiments}
Table \ref{tab:main} shows the overall results on the Multi-OpenEA benchmarks. 
\cjy{We have the following observations.
First, }
LODEME achieves \cjy{the best} performance on \cjy{all the} eight benchmarks.
\cjy{Although MSNEA and EVA are competitive, they are supervised while LODEME is unsupervised (self-supervised).
The good performance of LODEME can be explained in two aspects: 
its} PR module considers the holistic logical consistency \cjy{via} reasoning and provides SE module with high-quality \cjy{seed mappings}; 
\cjy{its SE module fully utilizes the images.}
%
Second, the four \cjy{modified} embedding-based models are \cjy{also} effective \cjy{in utilization of the} images, with an average rise \cjy{of 0.125 on Hit@1}. Note the highest rise on each benchmark is underlined. 
\cjy{BootEA-V, MultiKE-V and IMUSE-V achieve higher performance rise than RDGCN.}
%
\cjy{Third, the improvement due to the visual modality varies from KG to KG. 
Comparing V1 (by sparse KGs with lower average relation degree) and V2 (by dense KGs with higher average relation degree), the visual modality leads to an average rise of 0.143 and 0.106 on Hit@1, respectively.
This} indicates that images are ideal complementary information for long-tailed entities or sparse KGs with \cjy{limited} structural information.

\subsection{Ablation Experiments on Modalities}
\label{sec:ablation}
We report the results of the ablation experiments on \cjy{different} modalities in Table \ref{tab:ablations}, \cjy{where Name indicates entities names, Rel. \& Attr. indicates indicates relations \& attributes (including literals), w/o indicates not using a modality}. After removing the name information, the performance of EVA shows a severe decline compared to LODEME, indicating that EVA relies \cjy{more} on the name information, while LODEME adequately captures the visual information \cjy{leading to less reliance on the name information.}

In terms of \cjy{the importance of different modalities}, after dropping the structural information, Hit@1 of LODEME decreases by an average of 8.6\%, indicating that structural information is still the most important modality, which is consistent with the findings in \cite{liu2021visual}. Relatively, \cjy{the visual modality play a more important role}
than entity names, \cjy{and entity attributes \& relations. Hence, we recommend that future studies use more visual information and discard entity names, considering the \textit{name bias problems} also mentioned in many recent works \cite{liu2020exploring,sun2021knowing}.
Removing all (three) images causes a Hit@1 drop of 2.6\% and 3.9\% on two benchmarks, which are higher than the drop by removing entity names or entity relations \& attributes. 
Besides, we analyzed removing different numbers of images.}

\begin{table}[h]
\centering
\caption{\cjy{Results of a}blation experiments on modalities.}
\scalebox{0.75}{
\begin{tabular}{lcccccc}
\toprule
\multirow{2}{*}{Model}        & \multicolumn{3}{c}{D-W-100K-V1} & \multicolumn{3}{c}{EN-FR-100K-V1} \\ \cmidrule{2-7} 
                         & Hit@1      & Hit@5     & MRR     & Hit@1      & Hit@5      & MRR      \\ \midrule
EVA                      &  0.968          & 0.989       & 0.976        & 0.940           & 0.968           & 0.950         \\
w/o Name             & 0.822           & 0.910       & 0.854        & 0.712           & 0.836           & 0.752         \\ \midrule
LODEME              & 0.973           &    0.992       & 0.973        & 0.966            & 0.983           & 0.972         \\
w/o Structure     & 0.870           & 0.917           & 0.887        & 0.897           & 0.952          & 0.918         \\
w/o Name     & 0.967           & 0.986           & 0.968       & 0.931           & 0.962           & 0.942         \\
w/o Rel. \& Attr.     & 0.948           & 0.963           & 0.945        & 0.963           & 0.978           & 0.969        \\
w/o 1 Image & 0.963           & 0.978          & 0.968        &  0.956          & 0.972           & 0.962          \\
w/o 2 Images & 0.958           & 0.972           & 0.960        & 0.952           & 0.969           & 0.959         \\
w/o 3 Images & 0.947           & 0.967          & 0.953        & 0.927           & 0.968           & 0.948          \\ \bottomrule
\end{tabular}}
\vspace{-0.3cm}
\label{tab:ablations}
\end{table}

\subsection{Ablation Experiments on Use Strategy of Multiple Images}
In LODEME, we utilize the structure-aware attention mechanism to exploit the information of multiple images. To verify its effectiveness, we compare the operation of mean pooling of multiple images directly. Another stronger strategy is to retain only the image pairs with the highest similarity among the aligned pairs (there is information leakage due to the need to know the aligned pairs in advance).

The results are shown in Table \ref{tab:ablations2}, where the attention mechanism achieves better results compared to the mean pooling, proving its effectiveness. Also retaining the most similar image pairs achieves competitive results, which guides future work to investigate from the perspective of how to compute image similarity more accurately, such as similarity based on regions rather than complete images.

\begin{table}[h]
\centering
\caption{Ablation experiments on use strategy of multiple images.}
\scalebox{0.75}{
\begin{tabular}{lcccccc}
\toprule
\multirow{2}{*}{Strategy}        & \multicolumn{3}{c}{D-W-100K-V1} & \multicolumn{3}{c}{EN-FR-100K-V1} \\ \cmidrule{2-7} 
                         & Hit@1      & Hit@5     & MRR     & Hit@1      & Hit@5      & MRR      \\ \midrule
Attention              & 0.973           &    0.992       & 0.973        & 0.966            & 0.983           & 0.972         \\
Mean     & 0.954           & 0.962           & 0.958        & 0.942           & 0.966          & 0.957         \\
Highest Similarity & 0.970           & 0.994          & 0.965        & 0.960           & 0.979           & 0.964          \\ \bottomrule
\end{tabular}}
\vspace{-0.3cm}
\label{tab:ablations2}
\end{table}

\subsection{Overall Results On Existing Benchmarks}
We conducted experiments on the existing multi-modal EA benchmarks, and since images of DB15K are not public for the FB15K-DB15K dataset in Table \ref{tab:dataset_compare}, we conducted experiments on the DBP-WD(norm) dataset only, as shown in Table \ref{table:existingbenchmark}. LODEME also achieved state-of-the-art results.
However, due to the low image coverage, the visual information of the DBP-WD(norm) dataset does not bring as much gain as Multi-OpenEA for the four embedding-based models.

\begin{table}[h]
\centering
\renewcommand\arraystretch{1}
\caption{Overall results on existing benchmarks.}
\scalebox{0.7}{
\begin{tabular}{|l|ccc|}
\hline
\rowcolor[HTML]{EFEFEF} 
\cellcolor[HTML]{EFEFEF}                   & \multicolumn{3}{c|}{\cellcolor[HTML]{EFEFEF}DBP-WD(norm)}  \\
\cline{2-4} 
\rowcolor[HTML]{EFEFEF} 
\multirow{-2}{*}{\cellcolor[HTML]{EFEFEF}} & Hit@1           & Hit@5           & MRR                   \\ \hline
BootEA                                     & 0.323            & -            & 0.420                     \\
BootEA-V                                     & 0.362            & 0.581            & 0.474                      \\
MultiKE                                     & 0.096            & 0.210            & 0.159                      \\
MultiKE-V                                     & 0.187            & 0.346            & 0.257                      \\
RDGCN                                     & 0.138            & 0.268            & 0.203                     \\
RDGCN-V                                     & 0.169            & 0.312            & 0.238                    \\
IMUSE                                     & 0.104            & 0.231            & 0.174                    \\
IMUSE-V                                     & 0.143            & 0.269            & 0.211                     \\ \cline{1-4}
PARIS                                     & 0.834            & -            & -                    \\
EVA                                     & 0.985            & -            & 0.989                    \\
LODEME                                     & 0.991            & 0.998            & 0.992                 \\\hline
\end{tabular}}
\label{table:existingbenchmark}
\vspace{-0.3cm}
\end{table}

\section{Conclusion and Discussion}
In this study \cjy{we constructed eight large-scale EA benchmarks with multi-modal KGs, and 
evaluated four typical embedding-based entity alignment models which were extended for incorporating entity images.
We further proposed a new multi-modal EA method named LODEME using logical deduction and multi-modal KG embeddings.
The evaluation shows that the visual modality is \jy{always} quite effective 
in all these methods
while LODEME always achieves the best performance.}

Images have great potential to \cjy{further augment \jy{the EA methods.}
We show the results of some ablation experiments in}
Section \ref{sec:ablation}, 
\cjy{where}
we find that \cjy{the visual modality has a more positive impact on aligning more sparse KGs with weaker structure information. 
This motivates us to consider different image embedding and utilization solutions for KGs with different sparsities in the future. Meanwhile, we recommend that future studies use more visual information and discard entity names, considering the name bias. LODEME doesn't not consider fine-grained semantic types of the image objects, which could be utilized to avoid image noise. We leave this to be explored in future work.

\textbf{Acknowledge: }This research is supported by  National Natural Science Foundation of China (Grant No.62276154), AMiner.Shenzhen SciBrain Fund, Research  Center for Computer Network (Shenzhen) Ministry of Education and The EPSRC Project ConCur (EP/V050869/1).
}

\bibliographystyle{IEEEbib}
\bibliography{strings,refs}

\end{document}